\documentclass[12pt]{amsart}
\usepackage[table]{xcolor}
\usepackage{times}
\usepackage{amssymb, amsmath}
\usepackage{amsthm}
\usepackage{tikz}
\usetikzlibrary{matrix,arrows}
\usepackage{enumitem}
\usepackage{booktabs}
\usepackage{color}

\theoremstyle{plain}

\theoremstyle{remark}

\newcommand{\bs}{\boldsymbol}
\definecolor{Gray}{gray}{0.95}
\newcolumntype{g}{>{\columncolor{Gray}}c}

\begin{document}

\title[KDE sampling]{Kernel density estimation based sampling for imbalanced class distribution}
\author[F. Kamalov]{Firuz Kamalov}
\address{Canadian University Dubai, Dubai, UAE.}
\email{\textcolor[rgb]{0.00,0.00,0.84}{firuz@cud.ac.ae}}
\address{}
\email{}
\date{}

\begin{abstract}
Imbalanced response variable distribution is a common occurrence in data science. In fields such as fraud detection, medical diagnostics, system intrusion detection and many others where abnormal behavior is rarely observed the data under study often features disproportionate target class distribution. One common way to combat class imbalance is through resampling the minority class to achieve a more balanced distribution. In this paper, we investigate the performance of  the sampling method based on kernel density estimation (KDE). We believe that KDE offers a more natural way of generating new instances of minority class that is less prone to overfitting than other standard sampling techniques. It is based on a well established theory of nonparametric statistical estimation. Numerical experiments show that KDE can outperform other sampling techniques on a range of real life datasets as measured by $F_1$-score and G-mean. The results remain consistent across a number of classification algorithms used in the experiments. Furthermore, the proposed method outperforms the benchmark methods irregardless of the class distribution ratio. We conclude, based on the solid theoretical foundation and strong experimental results, that the proposed method would be a valuable tool in problems involving imbalanced class distribution.
\end{abstract}

\maketitle

\section{Introduction}
Imbalanced class distribution is a challenge that arises in many real world applications. 
It is prevalent in the fields of medical diagnostics, fraud detection, network intrusion detection and many others involving rare events \cite{krawcz}.   
Imbalanced class distribution usually appears in the context of a binary classification problem, where the response (target) variable consists of two classes with members of the majority class vastly outnumbering the members of the minority class. The majority and minority class members are labeled as negative and positive respectively.
In such cases, learning models tend to be biased in favor of the negatively labeled class. Concretely, a classifier trying to minimize its prediction error rate would focus more on the negatively labeled instances as they  comprise the majority of the data. At the same time, the positively labeled instances are often of more importance. For example, in medical diagnostics it would be critical to correctly identify patients with cancer though there may be relatively few such cases among the general population. 

To combat the problem of class imbalance researchers have proposed various strategies \cite{haix, lema} that can generally be divided into four categories: resampling, cost-sensitive learning, one class learning, and feature selection. Resampling involves balancing the class distribution by either undersampling the majority class or oversampling the minority class. This is a very popular approach that has been shown to perform well in various scenarios \cite{jian, maimon}.  
However, it is not without its limitations. Undersampling may lead to loss of potentially valuable information while oversampling may lead to overfitting. Cost sensitive learning is based on the idea of increasing the penalty for misclassifying the minority class instances. Since a classifier's objective is to minimize the overall cost there will be greater emphasis put on instances of minority class \cite{he, khan}. 
One class learning involves training a classifier on data with the target variable restricted to a single class. By ignoring all the majority class examples the classifier can get a clearer picture about the minority class \cite{raskutti}. 
Feature selection methods attempt to identify features that are effective in identifying minority class instances. This approach is particularly effective in cases involving high dimensional datasets \cite{haix2, maldonado}. 
Ensemble techniques that combine two or more techniques described above have also been explored in the literature. One common approach is to combine feature selection and oversampling to handle imbalanced data \cite{cao, triguero}.

In this paper, we propose a sampling approach based on kernel density estimation to deal with imbalanced class distribution. 
Kernel density estimation is a well-known method for estimating the unknown probability density distribution based on a given sample \cite{scott, silverman}. 
It estimates the unknown density function by averaging over a set of kernel homogeneous functions that are centered at each sample point. After having estimated the density distribution of the minority class we can then generate new sample points based on the density function. The proposed technique offers an intelligent and effective approach to synthesize new instances based on well-grounded statistical theory. Numerical experiments show that our method can perform better than other existing resampling techniques such as random sampling, SMOTE \cite{chawla, mathew}, ADASYN \cite{nguyen}, and NearMiss  \cite{mani}.  
The paper is organized as follows. In Section 2, we give an overview of the relevant literature for our study. In Section 3, we describe the methodology used in the study. We present our results in Section 4, and Section 5 concludes the paper.


\section{Literature}
The problem of class imbalance arises in a number of real-life applications and various approaches to address this issue have been put forth by researchers. 
Krawczyk \cite{krawcz} presents a good overview of the current trends in the field. One of the common ways to tackle class imbalance is through resampling whereby the majority class is undersampled and/or the minority class is oversampled. 
In the former, a portion of the majority class instances is sampled according to some strategy to achieve a more balanced class distribution.
Similarly, in the later approach the minority class is repeatedly sampled to increase its proportion relative to the majority class. 
One of the more popular undesampling techniques is NearMiss \cite{mani}, where the negative samples are selected so that the average distance to the $k$ closest samples of the positive class is the smallest. In a slightly different variation of NearMiss those negative samples are selected for which the average distance to the $k$ farthest samples of the positive class is the smallest. As shown by  Liu et al. \cite{lin, liu} an informed undersampling technique can lead to good results. However, in general, undesampling inevitably leads to loss of information. 
On the other hand, random sampling of the minority class (with replacement) can also cause issues such as overfitting \cite{chawla}. 
More advanced sampling techniques attempt to overcome the issue of overfitting by generating new samples of the minority class in a more intelligent manner. In this regard, Chawla et al. \cite{chawla} proposed a popular oversampling technique called SMOTE. In their approach, new instances are generated by random linear interpolation between the existing minority samples. Given a minority sample point $P_0$ a new random point is chosen along the line segment joining $P_0$ to one of its nearest neighbors $P_i$. This method has proven to be effective in a number of applications \cite{Fernandez}.  Another popular variant of SMOTE is an adaptive algorithm called ADASYN \cite{he2, nguyen}. It creates more examples in the neighborhood of the boundary between the two classes than in the interior of the minority class. There also exist nonlinear variants of SMOTE such as \cite{mathew}, where the authors interpolate between the points in the feature space of kernel SVM. 
An oversampling method based on Mahalanobis distance was proposed in \cite{abdi}. The authors propose to generate new samples of the minority class that have the same Mahalanobis distance from the class mean as the existing minority points.  Experiments showed that the method performs well on multi-class imbalanced data.
An intelligent combination of undersampling and oversampling techniques may also be an interesting avenue of study as was argued by the authors in \cite{jian}.

The sampling technique proposed in this paper relies on approximating the underlying density distribution of the minority class based on the existing samples. Probability density estimation techniques can be categorized into two groups: parametric and nonparametric. In parametric methods a fixed density function is assumed and its parameters are then estimated by maximizing the likelihood of obtaining the current sample. This approach introduces a specification bias and is susceptible to ovefitting \cite{lehmann}. 
Nonparametric approaches estimate the density distribution directly from the data. Among the nonparametric methods kernel density estimation (KDE) is the most popular approach in the literature \cite{silverman, simonoff}. 
It is a well established technique both within the statistical and machine learning communities \cite{botev, kim}. 
KDE has been successfully used in a wide array of applications including breast cancer data analysis \cite{sheikh}, 
image annotation \cite{yavlinky},
wind power forecast \cite{jeon},
and outlier detection \cite{kamalov}. 

A KDE based sampling approach was used in \cite{gao}, where the authors applied a 2-step procedure by first oversampling the minority samples using KDE and then constructing a radial basis function classifier. Numerical experiments on 6 datasets showed that their method performed better than comparable techniques. Our paper differs from \cite{gao} in that we perform a  systematic study of the KDE method. We delve deeper to analyze the differences between KDE and other sampling techniques. We  carry out a large number of numerical experiments to compare the performance of KDE  to other standard sampling methods.


\section{KDE sampling}
Nonparametric density estimation is an important tool in statistical data analysis. It is used to model the distribution of a variable based on a random sample. The resulting density function can be utilized to investigate various properties of the variable.  
Let $\{x_1, x_2, ..., x_n\}$ be an i.i.d. sample drawn from an unknown probability density function $f$. Then the kernel density estimate of $f$ is given by 

\begin{equation}
\tilde{f}(x) = \frac{1}{n}\sum_{i=1}^n K_h(x-x_i) ,
\end{equation}
where $K$ is the kernel function, $h$ is the bandwidth parameter, and $K_h(t) = \frac{1}{h}K(\frac{t}{h})$.  Intuitively, the true value of  $f(x)$ is estimated as the average distance from $x$ to the sample data points $x_i$. The 'distance' between $x$ and $x_i$ is calculated via a kernel function $K(t)$. There exists a number of kernel functions that can be used for this purpose including Epanechnikov, exponential, tophat, linear, and cosine. However, the most popular kernel  is the Gaussian function i.e.  
\begin{equation}
K(t) = \phi(t),
\end{equation}
where $\phi$ is the standard normal density distribution. The bandwidth parameter $h$
controls the smoothness of the density function estimate as well as the tradeoff between the bias and variance. A large value of $h$ results in a very smooth (low variance), but high bias density distribution. A small value of $h$ leads to an unsmooth (high variance), but low bias density distribution. The value of $h$ has a much bigger effect on the KDE approximation than the actual kernel. The value of $h$ can be determined by optimizing the mean integrated square error:
\begin{equation}
\mbox{MISE}(h) = E\big[\int (\tilde{f}(x)-f(x))^2\, dx\big].
\end{equation}
The MISE formula cannot be used directly since it involves the unknown density function $f$. Therefore, a number of other methods have been developed to determine the optimal value of $h$. The two most frequently used approaches to select the bandwidth value are rule of thumb methods and cross-validation. The rule of thumb methods approximate the optimal value of $h$ under certain assumptions about the underlying density function $f$ and its estimate $\tilde{f}$. A common approach is to use Scott's rule of thumb \cite{scott} for the value of $h$:
\begin{equation}\label{h_val}
h  = n^{-\frac{1}{5}}\cdot s,
\end{equation}
where $s$ is the sample  standard deviation. The optimal bandwidth value can also be determined numerically through cross-validation. It is done by applying a grid search method to find the value of $h$ that minimizes the sample mean integrated square error:
\begin{equation}
\mbox{MISE}_n(h) = \frac{1}{n} \sum_{i=1}^n (\tilde{f}(x_i)-f(x_i))^2.
\end{equation}

Kernel density estimation for multivariate variables follows essentially the same approach as the one dimensional approach described above. Given a sample $\{\bs{x}_1, \bs{x}_2, ... \bs{x}_n\}$ of $d$-dimensional random sample vectors drawn from a  distribution described by a density function $f$ the kernel density estimate is defined to be 

\begin{equation}\label{Kernel}
\bs{\tilde{f}_H(x)} = \frac{1}{n} \sum_{i=1}^n \bs{K_H (x-x_i)},
\end{equation}
where $\bs{H}$ is a $d\times d$ bandwidth matrix. The bandwidth matrix can be chosen in a variety of ways. In this study, we  use the multivariate version of Scott's rule:
\begin{equation}\label{H_val}
\bs{H}  = n^{-\frac{1}{d+4}}\cdot \bs{S},
\end{equation}
where $\bs{S}$ is a the data covariance matrix. Furthermore, we use  multivariate normal distribution as the kernel function:
\begin{equation}\label{K}
\bs{K_H(x)} = \frac{1}{\sqrt{(2\pi)^d |\bs{H}|}}e^{-\frac{1}{2}\bs{x^T H^{-1}x}}.
\end{equation} 

\begin{figure}[h!]
\centering
\includegraphics[scale=0.5]{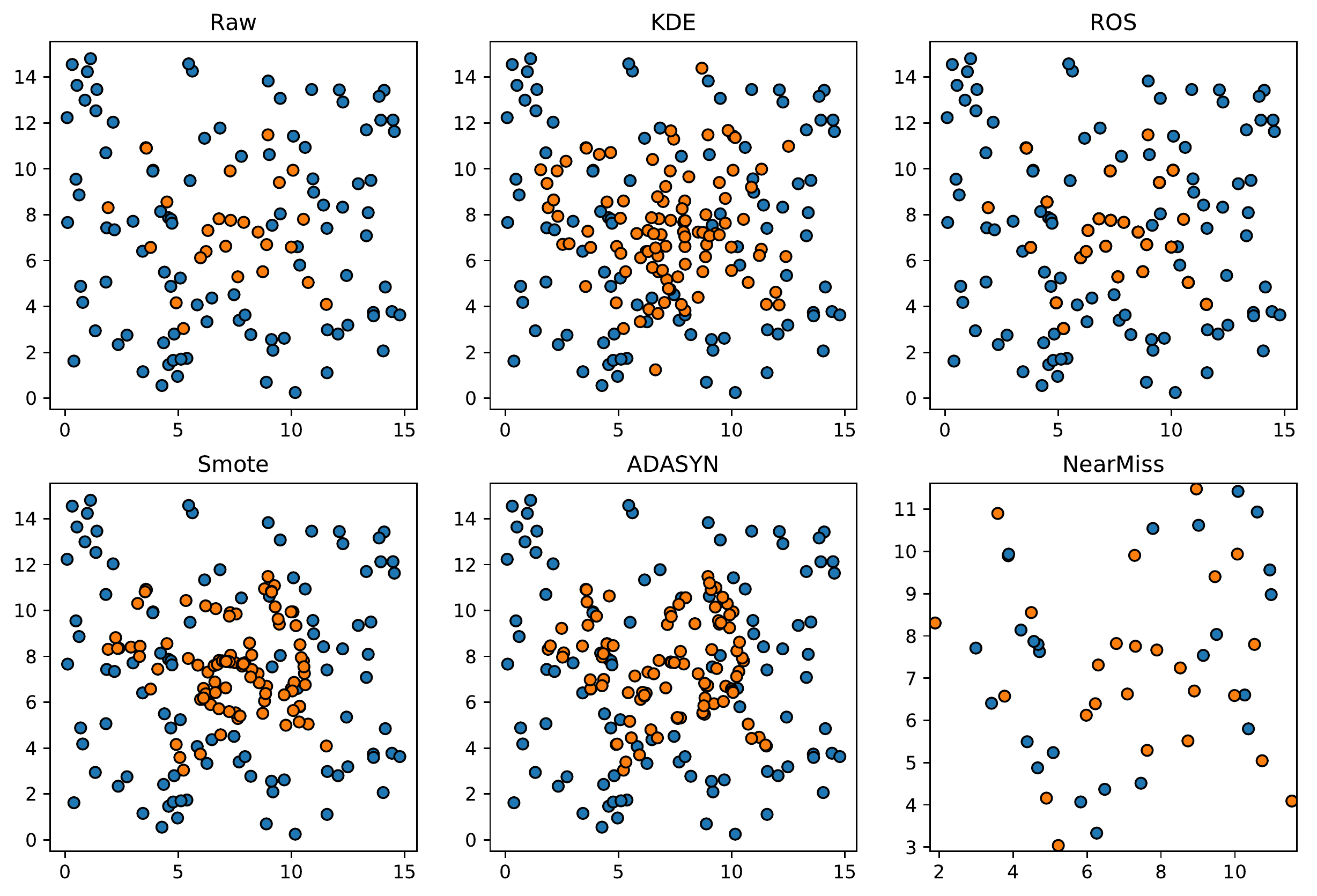}
\caption{Resampled data based on various techniques.}
\label{sample}
\end{figure}

The difference between KDE sampling and other standard sampling methods is illustrated in Figure \ref{sample}. The original data in the figure consists of 100 uniformly distributed blue points with the points in radius of 2 from the center being dropped. The 25 orange points are generated in the center of the figure via Gaussian distribution with standard deviation of 2. As can be seen from the figure, KDE creates new sample points by 'spraying' around the existing minority class points. The points are created using Gaussian distribution centered at randomly chosen existing minority class points. This process seems more intuitive than other sampling methods. On other hand, SMOTE method creates new sample points by interpolating between the existing minority class points. As a result all SMOTE generated points lie in the convex hull of the original minority class samples. Therefore, the new sampled data does not represent well the true underlying population distribution.  Random sampling with replacement (ROS) creates new points by simply resampling the existing minority class points. As a result the new sampled data  differs very little from the original data though it is  more dense at each sample location. ADASYN sampled plot resembles the SMOTE plot but  creates a bigger number of points at the edge of the minority cluster. NearMiss undersamples from the majority class thereby losing a lot of information as can be seen from its plot.

\section{Numerical Experiments}
In this section, we carry out a number of experiments to evaluate the performance of the KDE sampling method. To this end, we compare KDE to four standard sampling approaches used in the literature: Random Oversampling (ROS), SMOTE, ADASYN, and NearMiss. 
The ROS method is the simplest approach to generate additional minority class instances. Concretely, the new instances are randomly selected from the minority class with replacement. The ROS method serves as a basic benchmark for other sampling methods. 
The SMOTE algorithm is a widely used oversampling technique for generating new instances of the minority class \cite{chawla}. Unlike ROS, the SMOTE algorithm creates new instances synthetically. Concretely, for each point in the minority class its $k$ nearest neighbors in the minority class are determined. Given a minority point and one of its neighbors, the difference between the two is calculated and multiplied by a random number between $0$ and $1$. The new minority class instance is obtained by adding the preceding result to the minority point:
\begin{equation}\label{smote}
p^-_{new} = p^- + t(p^- - p^-_k),
\end{equation}
where $p^-_{new}$ denotes the new minority class point, $p^-$ is the minority point under consideration, $p^-_k$ is a $k$th nearest neighbor, and $t$ is a random number between $0$ and $1$. Thus, the SMOTE algorithm creates new instances by linearly interpolating between $k$ nearest neighbors of minority class points. 
ADASYN  is a popular extension of SMOTE \cite{he2}. It uses the same algorithm as SMOTE (Equation \ref{smote}) to generate new instances of the minority class. However, ADASYN generates more points around the minority instances which are closer to the majority class. In particular, for each minority instance $p^-$, its $k$ nearest neighbors are determined and the learning difficulty rate is calculated according to: 
\begin{equation}
r = \frac{m}{k},
\end{equation}  
where $m$ is the number of majority class members in the $k$ nearest neighborhood. Consequently, ADASYN generates more points around the minority instances with higher $r$ values.
NearMiss is a popular undersampling technique used to reduce the size of the majority class data \cite{mani}. It aims to sample instances of the majority class in a manner that preserves the original boundary structure of the data. There exist several variations of the NearMiss algorithm. In the most popular version the NearMiss algorithm, the majority class instances that are closest to the minority class are selected. To this end, pairwise distances between members of the majority and minority classes are calculated. Then for each point in the majority class $p^+$, the average distance to the closest $k$ points of the minority class $\overline{d}_{p^+}$ is calculated. The points $p^+$ with the corresponding smallest values of $\overline{d}_{p^+}$ are selected as the majority class sample.

The implementations of all four sampling approaches are taken from the \emph{imblearn} Python library \cite{lema} with their default settings. The implementation of KDE is taken from \emph{scipy.stats} Python library \cite{jones} with its default settings. In particular, we used the multivariate Gaussian KDE with its default bandwidth value determined by Scott's Rule (see Equations \ref{Kernel} - \ref{K}). Note that the performance of the KDE method can be further optimized by choosing the bandwidth value via cross validation.

The usual measures of classifier performance such as the accuracy rate are not suitable in the context of imbalanced datasets because their results can be misleading. For instance, given a dataset with 90$\%$ of instances labeled as negative, we can achieve a 90$\%$ accuracy rate by simply guessing all the instances to be negative. 
Although we would obtain a high accuracy rate we would miss all the positive instances in the data.
Ideally, we would like a metric that would measure classifier performance on both classes. 
To address this issue, authors often use the area under receiver operating characteristic curve (AUC) \cite{chawla, moy}. AUC reflects classifier performance based on true positive and false positive rates and it is not sensitive to class imbalance  \cite{fawcett}.
However, calculating AUC requires probabilities of the predicted labels which are not readily available for certain algorithms including KNN and SVM. Therefore, as an alternative to AUC, we also use G-mean: 
\begin{equation}
\mbox{G-mean} = \sqrt{\frac{TP}{TP+FN}\cdot \frac{TN}{TN+FP}}
\end{equation}
and $F_1$-score:
\begin{equation}
F_1 =2\cdot\frac{precision\cdot recall}{precision + recall}
\end{equation}
to measure classifier performance \cite{maldonado, moy}.


\subsection{Simulated Data}
\begin{figure}[h!]
\centering
\includegraphics[scale=0.4]{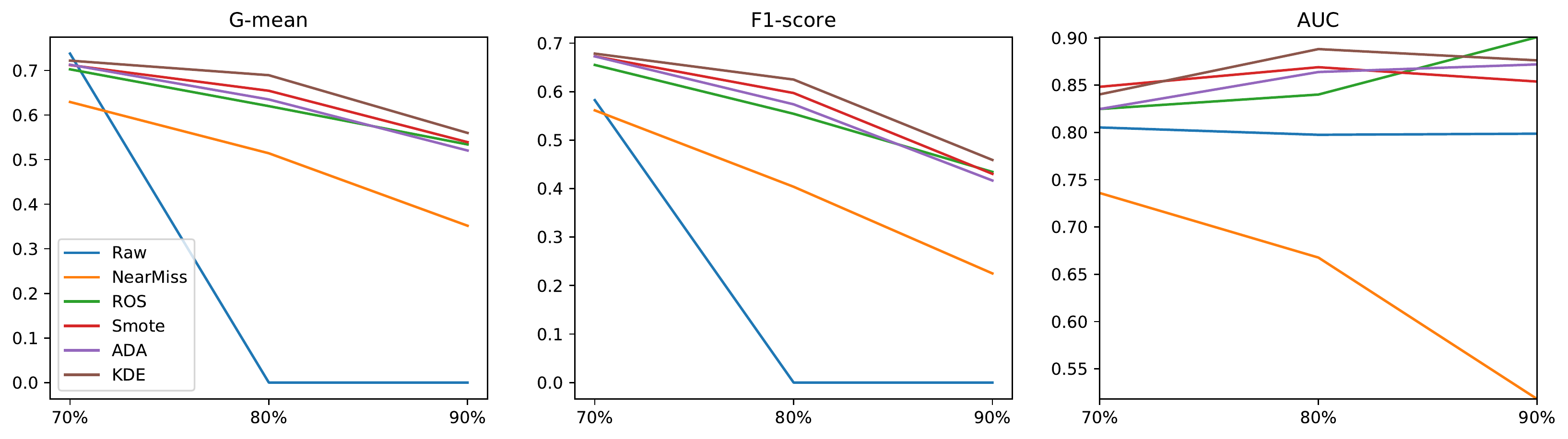}
\caption{Performance of sampling  method under varying class imbalance ratios.}
\label{sample_exp}
\end{figure}

We begin by considering a situation similar to the one described in Figure \ref{sample}. 
We use a dataset of size 1000 where the majority class points are uniformly distributed over a square grid. Next, we remove the points within radius of 2 from the center from the majority set. The minority class points are simulated using  Gaussian distribution with mean at the center of the grid and standard deviation of 2. We measure the performance of the sampling methods over a range of class imbalance ratios. We use a feedforward neural network with one hidden layer as our base classifier. The results of the experiment are presented in Figure \ref{sample_exp}. We see that the KDE sampling technique outperforms other methods as measured by G-mean and $F_1$-score. Moreover, KDE has the edge under different class imbalance ratios. As measured by AUC, KDE is the best at 80$\%$ imbalance  ratio and second best at $70\%$ and $90\%$ imbalance ratios.  

\begin{figure}[h!]
\centering
\includegraphics[scale=0.4]{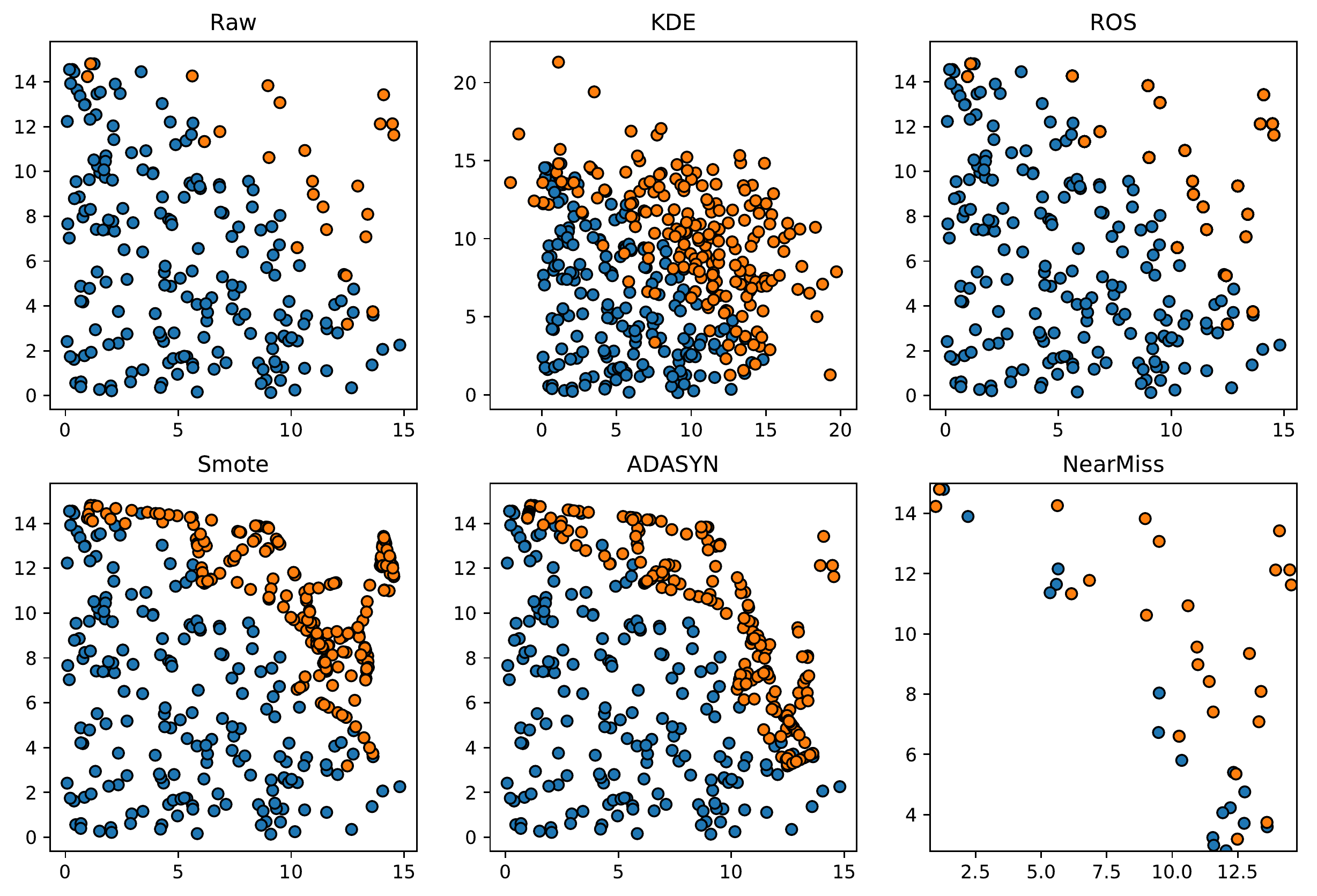}
\caption{Resampled data for a nearly (linearly) separable data.}
\label{sample_exp2}
\end{figure}

Next, we consider a dataset consisting of 500 majority and 100 minority class instances. The sample points are generated using a uniform distribution. The data is constructed to be almost linearly separable (see Figure  \ref{sample_exp2}a). We apply various sampling techniques to the raw data with the results illustrated in Figures \ref{sample_exp2}b-\ref{sample_exp2}f.
As can be seen from the figures the new KDE minority samples are spread across a larger region. On the other hand, ROS, SMOTE, and ADASYN generated samples are more concentrated making them more prone to overfitting.  

A feedfoward neural network is trained on each resampled dataset. The AUC results are presented in Table \ref{tab4}. As can be seen from the table, KDE significantly outperforms the other sampling techniques. 

\begin{table}[h!]
\centering
\caption{AUC results for data in Figure \ref{sample_exp2}.}
\label{tab4}
\begin{tabular}{lrrrrrr}
\toprule
{Metric} &  Raw &  NearMiss &  ROS &  SMOTE &    ADASYN &    KDE \\
\midrule
AUC      &  0.757 &   0.727 &   0.820 &  0.8301 &  0.814 &  0.871 \\
\bottomrule
\end{tabular}
\end{table}

\begin{figure}[h!]
\centering
\includegraphics[scale=0.4]{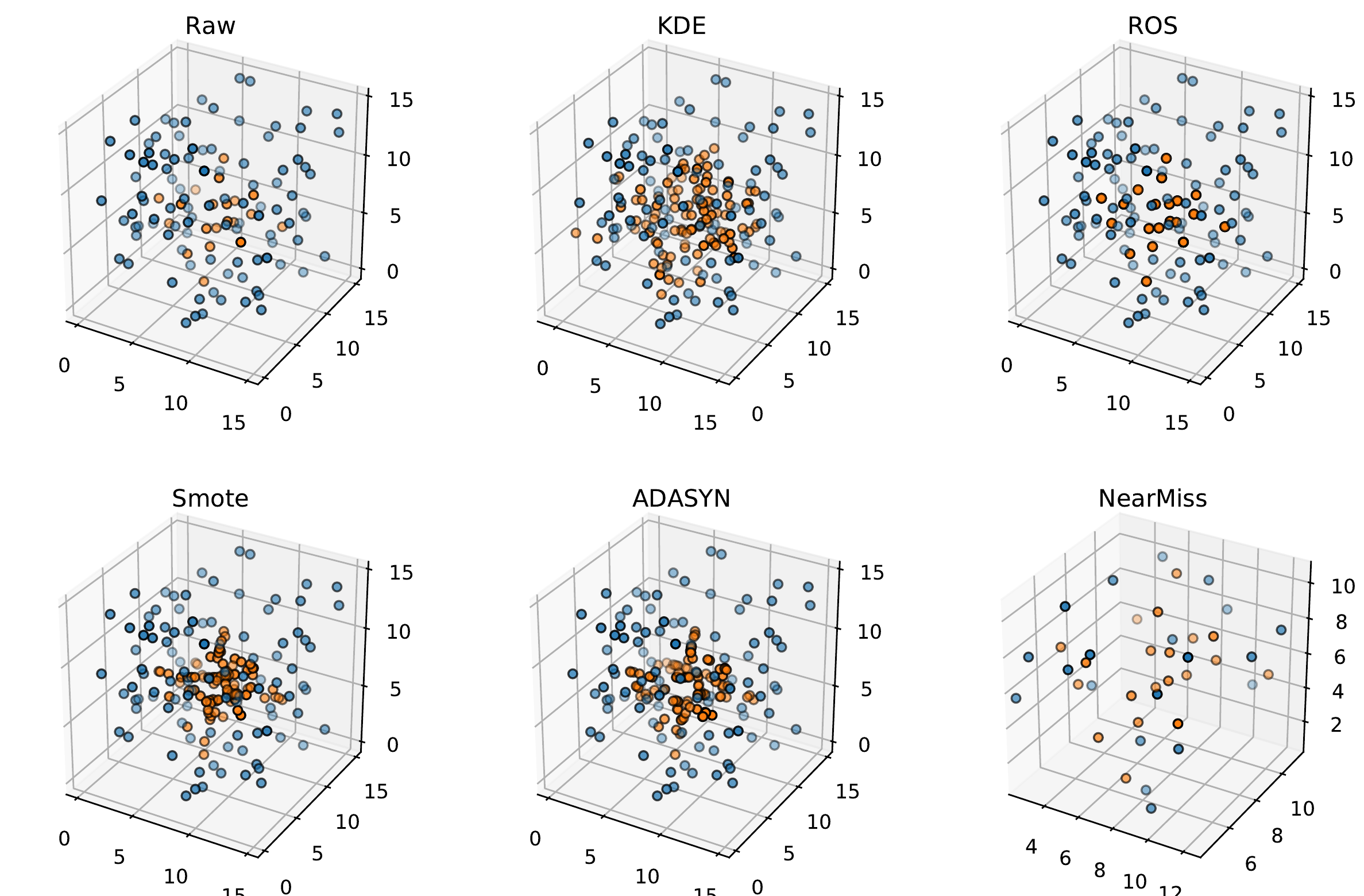}
\caption{Generated samples in 3D.}
\label{3d}
\end{figure}

Our last illustration is in three-dimensional space as shown in Figure \ref{3d}. The majority class samples consist of 500 uniformly distributed points over the cube $[0, 15]^{3}$ with the sphere of radius 1.5 removed from the center of the set. The minority class samples consist of 100 points generated according to the Gaussian distribution with $\mu = (7, 7, 7)$ and $\sigma = 2$. As can be seen from Figure \ref{3d}, the KDE resampled data appears to be more diffused whereas ROS, SMOTE, and ADASYN generated data is more concentrated. 

A feedfoward neural network is trained on each resampled dataset and the results are presented in Table \ref{tab_3d}. As can be observed from the table, KDE achieves the best results in AUC and $F_1$-score. And it is second best in terms of G-mean. 

\begin{table}[h!]
\centering
\caption{AUC results for data in Figure \ref{3d}.}
\label{tab_3d}
\begin{tabular}{lllllll}
\toprule
{} &       Raw &  NearMiss &    ROS &     SMOTE &       ADASYN &       KDE \\
\midrule
AUC     &  0.870593 &   0.74237 &  0.883333 &  0.874519 &  0.862889 &  0.890148 \\
G-mean   &   0.17598 &  0.554593 &   0.63151 &  0.602904 &  0.610252 &  0.618056 \\
$F_1$-score &  0.020202 &  0.403148 &  0.544493 &  0.511166 &  0.519447 &  0.546625 \\
\bottomrule
\end{tabular}
\end{table}


\subsection{Real Life Data}
In order to obtain a comprehensive evaluation of our method we used a range of datasets and classifiers. In particular, we used 12 real life datasets with various class imbalance ratios ranging from $1.86:1$ to $42:1$ (Table \ref{info}). Each sampling method is tested on 3 separate base classifiers: $k$-nearest neighbors (KNN), support vector machines (SVM), and multilayer perceptron (MLP).

\begin{table}[h!]
\centering
\caption{Experimental Datasets}
\label{info}
\begin{tabular}{llllll}
\toprule
{} &          Name &         Repository \& Target &   Ratio &      \#S &   \#F \\
\midrule
\rowcolor{Gray}
1  &      diabetes &              UCI, target: 1 &  1.86:1 &     768 &    8 \\
2  &          bank &            UCI, target: yes &   7.6:1 &  43,193 &   24 \\
\rowcolor{Gray}
3  &         ecoli &            UCI, target: imU &   8.6:1 &     336 &    7 \\
4  &      satimage &              UCI, target: 4 &   9.3:1 &   6,435 &   36 \\
\rowcolor{Gray}
5  &       abalone &              UCI, target: 7 &   9.7:1 &   4,177 &   10 \\
6  &  spectrometer &           UCI, target: $>$=44 &    11:1 &     531 &   93 \\
\rowcolor{Gray}
7  &     yeast\_ml8 &           LIBSVM, target: 8 &    13:1 &   2,417 &  103 \\
8  &         scene &  LIBSVM, target: $>$one label &    13:1 &   2,407 &  294 \\
\rowcolor{Gray}
9  &   libras\_move &              UCI, target: 1 &    14:1 &     360 &   90 \\
10 &  wine\_quality &      UCI, wine, target: $<$=4 &    26:1 &   4,898 &   11 \\
\rowcolor{Gray}
11 &    letter\_img &              UCI, target: Z &    26:1 &  20,000 &   16 \\
12 &     yeast\_me2 &            UCI, target: ME2 &    28:1 &   1,484 &    8 \\
\rowcolor{Gray}
13 &   ozone\_level &            UCI, ozone, data &    34:1 &   2,536 &   72 \\
14 &   mammography &       UCI, target: minority &    42:1 &  11,183 &    6 \\
\bottomrule
\end{tabular}
\end{table}

During the experiments the data was split into training and testing parts. The results based on the testing part are calculated and reported in the study. Furthermore, each experiment was run twice using different training/testing splits. The average value of the results of the two runs are being presented in the paper.
The results for each classifier are summarized in 3 separate tables below. When using the KNN algorithm, the KDE sampling method often yields significantly better results compared to other sampling methods (see Table \ref{knn}). For instance, when used on \emph{ecoli} dataset the KDE method produces G-mean of 0.753 which is 5$\%$ better than the second best method (SMOTE); and $F_1$-score of 0.691 which is 6$\%$ better than the second best method. Note that the KDE method performs well on datasets with both low and high imbalance ratios. 

\begin{table}[h!]
\centering
\caption{G-mean and $F_1$-score results based on KNN classifier.}
\label{knn}

\begin{tabular}{lrrrrr}
\toprule
{} &  NearMiss &  ROS &  SMOTE &    ADASYN &    KDE \\
\midrule
\rowcolor{Gray}
diabetes G      &     0.685 &   0.693 &  0.695 &  0.673 &  \textbf{0.711} \\
diabetes $F_1$     &     0.613 &   0.623 &  0.634 &  0.614 &  0.622 \\
\rowcolor{Gray}
bank G          &     0.393 &   0.584 &  0.584 &  0.575 &  \textbf{0.689} \\
bank $F_1$         &     0.270 &   0.461 &  0.470 &  0.462 &  0.345 \\
\rowcolor{Gray}
ecoli G         &     0.457 &   0.679 &  0.705 &  0.686 &  \textbf{0.753} \\
ecoli $F_1$        &     0.339 &   0.593 &  0.631 &  0.611 &  \textbf{0.691} \\
\rowcolor{Gray}
satimage G      &     0.352 &   0.715 &  0.670 &  0.653 &  \textbf{0.732} \\
satimage $F_1$     &     0.223 &   0.650 &  0.608 &  0.589 &  0.633 \\
\rowcolor{Gray}
abalone G       &     0.200 &   0.478 &  0.470 &  0.478 &  \textbf{0.492} \\
abalone $F_1$      &     0.098 &   0.336 &  0.339 &  0.350 &  0.191 \\
\rowcolor{Gray}
spectrometer G  &     0.641 &   0.936 &  0.890 &  0.915 &  \textbf{0.961} \\
spectrometer $F_1$ &     0.521 &   0.771 &  0.754 &  0.771 &  0.723 \\
\rowcolor{Gray}
yeast\_ml8 G     &     0.316 &   0.284 &  0.294 &  0.298 &  \textbf{0.557} \\
yeast\_ml8 $F_1$    &     0.172 &   0.125 &  0.161 &  0.166 &  0.068 \\
\rowcolor{Gray}
scene G         &     0.302 &   0.451 &  0.372 &  0.368 &  \textbf{0.546} \\
scene $F_1$        &     0.145 &   0.285 &  0.241 &  0.236 &  0.096 \\
\rowcolor{Gray}
libras\_move G   &     0.842 &   0.823 &  0.760 &  0.740 &  \textbf{0.874} \\
libras\_move $F_1$  &     0.647 &   0.806 &  0.730 &  0.707 & \textbf{0.842} \\
\rowcolor{Gray}
wine\_quality G  &     0.270 &   0.456 &  0.393 &  0.395 &  \textbf{0.527} \\
wine\_quality $F_1$ &     0.129 &   0.291 &  0.252 &  0.256 &  \textbf{0.335} \\
\rowcolor{Gray}
letter\_img G    &     0.438 &   0.956 &  0.940 &  0.938 &  0.943 \\
letter\_img $F_1$   &     0.321 &   0.945 &  0.932 &  0.931 &  0.934 \\
\rowcolor{Gray}
yeast\_me2 G     &     0.300 &   0.458 &  0.488 &  0.475 &  0.465 \\
yeast\_me2 $F_1$    &     0.153 &   0.285 &  0.366 &  0.344 &  0.293 \\
\rowcolor{Gray}
ozone\_level G   &     0.134 &   0.390 &  0.335 &  0.348 &  0.374 \\
ozone\_level $F_1$  &     0.036 &   0.209 &  0.189 &  0.205 &  0.111 \\
\rowcolor{Gray}
mammography G   &     0.179 &   0.666 &  0.567 &  0.522 &  \textbf{0.663} \\
mammography $F_1$  &     0.062 &   0.570 &  0.469 &  0.416 &  \textbf{0.568} \\
\bottomrule
\end{tabular}
\end{table}

Applying SVM to compare the sampling methods produces results that are similar to KNN. As can be seen from Table \ref{svm}, KDE often yields significantly better results than other sampling methods.   For instance, when used on \emph{spectrometer} dataset the KDE method produces G-mean of 0.924 which is 12$\%$ better than the second best method (SMOTE); and $F_1$-score of 0.878 which is 14$\%$ better than the second best method. Note again that the KDE method performs well on datasets with both low and high imbalance ratios.

\begin{table}
\centering
\caption{G-mean and $F_1$-score results based on SVM classifier.}
\label{svm}
\begin{tabular}{lrrrrr}
\toprule
{} &  NearMiss &  ROS &  SMOTE &    ADASYN &    KDE \\
\midrule
\rowcolor{Gray}
diabetes G      &     0.681 &   0.705 &  0.701 &  0.704 &  \textbf{0.706} \\
diabetes $F_1$     &     0.626 &   0.647 &  0.633 &  0.654 &  0.635 \\
\rowcolor{Gray}
bank G          &     0.378 &   0.594 &  0.599 &  0.581 &  \textbf{0.701} \\
bank $F_1$         &     0.255 &   0.504 &  0.507 &  0.489 &  0.411 \\
\rowcolor{Gray}
ecoli G         &     0.309 &   0.699 &  0.698 &  0.698 &  \textbf{0.709} \\
ecoli $F_1$        &     0.190 &   0.648 &  0.636 &  0.636 &  \textbf{0.659} \\
\rowcolor{Gray}
satimage G      &     0.327 &   0.652 &  0.663 &  0.612 &  0.618 \\
satimage $F_1$     &     0.199 &   0.582 &  0.596 &  0.538 &  0.542 \\
\rowcolor{Gray}
abalone G       &     0.184 &   0.494 &  0.492 &  0.485 &  \textbf{0.508} \\
abalone $F_1$      &     0.085 &   0.389 &  0.385 &  0.377 &  \textbf{0.401} \\
\rowcolor{Gray}
spectrometer G  &     0.555 &   0.795 &  0.808 &  0.802 &  \textbf{0.924} \\
spectrometer $F_1$ &     0.436 &   0.720 &  0.732 &  0.738 &  \textbf{0.878} \\
\rowcolor{Gray}
yeast\_ml8 G     &     0.276 &   0.264 &  0.278 &  0.278 &  na \\
yeast\_ml8 $F_1$    &     0.146 &   0.048 &  0.018 &  0.018 &  na \\
\rowcolor{Gray}
scene G         &     0.279 &   0.622 &  0.583 &  0.578 &  0.472 \\
scene $F_1$        &     0.149 &   0.349 &  0.314 &  0.306 &  0.178 \\
\rowcolor{Gray}
libras\_move G   &     0.351 &   0.867 &  0.886 &  0.886 &  \textbf{0.935} \\
libras\_move $F_1$  &     0.218 &   0.804 &  0.878 &  0.878 &  \textbf{0.933} \\
\rowcolor{Gray}
wine\_quality G  &     0.235 &   0.404 &  0.413 &  0.405 &  \textbf{0.440} \\
wine\_quality $F_1$ &     0.105 &   0.261 &  0.271 &  0.263 &  \textbf{0.287} \\
\rowcolor{Gray}
letter\_img G    &     0.462 &   0.944 &  0.963 &  0.973 &  0.796 \\
letter\_img $F_1$   &     0.351 &   0.932 &  0.951 &  0.961 &  0.772 \\
\rowcolor{Gray}
yeast\_me2 G     &     0.217 &   0.430 &  0.456 &  0.443 &  \textbf{0.504} \\
yeast\_me2 $F_1$    &     0.089 &   0.293 &  0.323 &  0.309 &  \textbf{0.380} \\
\rowcolor{Gray}
ozone\_level G   &     0.146 &   0.446 &  0.451 &  0.436 &  0.426 \\
ozone\_level $F_1$  &     0.043 &   0.294 &  0.296 &  0.278 &  0.262 \\
\rowcolor{Gray}
mammography G   &     0.191 &   0.517 &  0.553 &  0.472 &  0.535 \\
mammography $F_1$  &     0.070 &   0.411 &  0.454 &  0.355 &  0.427 \\
\bottomrule
\end{tabular}
\end{table}

Using the MLP classifier does not produce as strong of results as using KNN or SVM. Although there are instances - \emph{ecoli, mammography} - where KDE outperforms other sampling methods its performance is not overwhelming (see Table \ref{nn}). This may be the result of the particular network architecture used in the experiment: a single hidden layer with 32 fully  connected nodes. It is possible that a different architecture may produce better results for KDE sampling. 

\begin{table}[h!]
\centering
\caption{G-mean and $F_1$-score results based on MLP classifier.}
\label{nn}
\begin{tabular}{lrrrrr}
\toprule
{} &  NearMiss &  ROS &  SMOTE &    ADASYN &    KDE \\
\midrule
\rowcolor{Gray}
diabetes G      &     0.712 &   0.725 &  0.715 &  0.702 &  0.695 \\
diabetes $F_1$     &     0.659 &   0.665 &  0.645 &  0.644 &  0.628 \\
\rowcolor{Gray}
bank G          &     0.388 &   0.607 &  0.614 &  0.589 &  \textbf{0.721} \\
bank $F_1$         &     0.266 &   0.519 &  0.525 &  0.498 &  0.377 \\

\rowcolor{Gray}
ecoli G         &     0.390 &   0.741 &  0.765 &  0.724 &  \textbf{0.762} \\
ecoli $F_1$        &     0.263 &   0.688 &  0.708 &  0.667 &  \textbf{0.724} \\
\rowcolor{Gray}
satimage G      &     0.337 &   0.722 &  0.734 &  0.761 &  0.655 \\
satimage $F_1$     &     0.209 &   0.648 &  0.652 &  0.674 &  0.555 \\
\rowcolor{Gray}
abalone G       &     0.197 &   0.513 &  0.513 &  0.498 &  \textbf{0.522} \\
abalone $F_1$      &     0.097 &   0.407 &  0.405 &  0.388 &  0.253 \\
\rowcolor{Gray}
spectrometer G  &     0.368 &   0.882 &  0.957 &  0.952 &  0.931 \\
spectrometer $F_1$ &     0.239 &   0.715 &  0.700 &  0.758 &  0.741 \\
\rowcolor{Gray}
yeast\_ml8 G     &     0.279 &   0.324 &  0.381 &  0.462 &  0.313 \\
yeast\_ml8 $F_1$    &     0.147 &   0.098 &  0.115 &  0.188 &  0.082 \\
\rowcolor{Gray}
scene G         &     0.311 &   0.537 &  0.504 &  0.516 &  0.466 \\
scene $F_1$       &     0.178 &   0.261 &  0.246 &  0.268 &  0.202 \\
\rowcolor{Gray}
libras\_move G   &     0.379 &   0.956 &  0.913 &  0.963 &  0.958 \\
libras\_move $F_1$  &     0.250 &   0.845 &  0.813 &  0.883 &  \textbf{0.890} \\
\rowcolor{Gray}
wine\_quality G  &     0.223 &   0.439 &  0.434 &  0.443 &  \textbf{0.449} \\
wine\_quality $F_1$ &     0.096 &   0.289 &  0.289 &  0.298 &  \textbf{0.299} \\
\rowcolor{Gray}
letter\_img G    &     0.593 &   0.980 &  0.971 &  0.971 &  0.882 \\
letter\_img $F_1$   &     0.517 &   0.964 &  0.954 &  0.948 &  0.873 \\
\rowcolor{Gray}
yeast\_me2 G     &     0.215 &   0.499 &  0.500 &  0.491 &  \textbf{0.623} \\
yeast\_me2 $F_1$    &     0.089 &   0.370 &  0.362 &  0.357 &  0.317 \\
\rowcolor{Gray}
ozone\_level G   &     0.178 &   0.493 &  0.444 &  0.364 &  0.406 \\
ozone\_level $F_1$  &     0.062 &   0.257 &  0.237 &  0.166 &  0.171 \\
\rowcolor{Gray}
mammography G   &     0.183 &   0.560 &  0.585 &  0.490 &  \textbf{0.629} \\
mammography $F_1$  &     0.065 &   0.467 &  0.497 &  0.378 &  \textbf{0.518} \\
\bottomrule
\end{tabular}
\end{table}


\section{Conclusion}
In this paper, we studied an oversampling technique based on KDE. We believe that KDE provides a natural and statistically sound approach to generating new minority samples in an imbalanced dataset. It allows creation of new instances with minimum overfitting. One of the main advantages of KDE technique is its flexibility. By choosing different kernel functions researchers can customize the sampling process. Additional flexibility is offered through selection of the  kernel bandwidth.  KDE is a well researched topic with an established statistical foundation. In addition, a variety of implementations of the KDE algorithm are available in Python, R, Julia and other programming languages. This makes KDE a very appealing tool to use in oversampling. 

We carried out a comprehensive study of the KDE sampling approach based on simulated and real life data. In particular, we used 3 simulated and 12 real life datasets that were tested on three different base classifiers. The proposed method was compared against four standard benchmark techniques for dealing with imbalanced class data. The results, as measured by $F_1$-score and G-mean, show that KDE has the ability to outperform other standard sampling methods in a number of different scenarios. The performance of KDE remains robust regardless of the base classifier used which indicates generalizability of the proposed method. Based on the above analysis, we conclude that KDE should be considered as a potent tool in dealing with the problem of imbalanced class distribution.


\end{document}